# A Mirroring Theorem and its application to a New method of Unsupervised Hierarchical Pattern Classification

Dasika Ratna Deepthi
Department of Computer Science,
Aurora's Engineering College, Bhongir,
Nalgonda Dist., A.P., India

K. Eswaran
Department of Computer Science,
Srinidhi Institute of Science and Technology,
Yamnampet, Ghatkesar, Hyderabad, India.

*Abstract—* **In this paper, we prove a crucial theorem called "Mirroring Theorem" which affirms that given a collection of samples with enough information in it such that it can be classified into classes and sub-classes then**

  **(i) There exists a mapping which classifies and sub-classifies these samples**

  **(ii) There exists a hierarchical classifier which can be constructed by using Mirroring Neural Networks (MNNs) in combination with a clustering algorithm that can approximate this mapping.**

**Thus, the proof of the Mirroring theorem provides a theoretical basis for the existence and a practical feasibility of constructing hierarchical classifiers, given the maps. Our proposed Mirroring Theorem can also be considered as an extension to Kolmogrov's theorem in providing a realistic solution for unsupervised classification. The techniques we develop, are general in nature and have led to the construction of learning machines which are (i) tree like in structure, (ii) modular (iii) with each module running on a common algorithm (tandem algorithm) and (iv) self-supervised. We have actually built the architecture, developed the tandem algorithm of such a hierarchical classifier and demonstrated it on an example problem.**

## I. INTRODUCTION

There have been various ways in which the fields of artificial intelligence and machine learning have been furthered: starting with experimentation [1], abstraction [2], [3] and the study of locomotion [4]. Many techniques have been developed to learn patterns [5] & [6] as well as to reduce large dimensional data [7] & [8] so that relevant information can be used for classification of patterns [9] & [10]. Investigators have tackled, to varying degrees of success, pattern recognition problems like face detection [11], gender classification [12], human expression recognition [13], object learning [14] & [15], unsupervised learning of new tasks [16] and also have studied complex neuronal properties of higher cortical areas [17], naming but a few. However, most of the above techniques did not require automatic feature extraction as a pre-processing step to pattern classification. In our approach, we developed a self-learning machine (based on our proposed Mirroring Theorem) which performs feature extraction and pattern learning simultaneously to recognize/classify the patterns in an unsupervised mode. This automatic feature extraction step, prior to unsupervised classification fulfills one more additional crucial requirement called dimensional reduction. Furthermore, by proving our stated mirroring theorem, we actually demonstrate that such unsupervised hierarchical classifiers mathematically exist. It is also proved that the hierarchical classifiers that do perform a level-by-level unsupervised classification can be approximated by a network of "nodes" forming a tree-like architecture. What we term as a "node", in this architecture, is actually an abstraction of an entity which executes two procedures: the "Mirroring Neural Network" (MNN) algorithm coupled with a clustering algorithm. The MNN performs automatic data reduction and feature extraction (see [18] for more details on MNN) and clustering does the subsequent step called unsupervised classification (of the extracted features of the MNN); these two steps are performed in tandem - hence our term Tandem Algorithm. The Mirroring Theorem provides a proof that this technique will always work provided sufficient information is contained in the ensemble of samples for it to be classified and sub-classified and certain continuity conditions for the mappings are satisfied. The Mirroring Theorem, we prove in this paper may be considered as an extension to Kolmogrov's theorem [19] in providing a practical method for unsupervised classification. The details of the theorem and how it can be used for developing an unsupervised hierarchical classifier are discussed in the next sections.

Our main contribution in this paper, is that we propose and prove a theorem called "Mirroring Theorem" which provides a mathematical base for constructing a new kind of architecture that performs an unsupervised hierarchical classification of its inputs by implementing a single common algorithm (which we call as the "Tandem Algorithm") and this is demonstrated on an example set of image patterns. That is, the proposed hierarchical classifier is mathematically proved to exist, for which we develop a new common algorithm that does the two machine steps, namely, automatic feature extraction and clustering to execute a level-by-level unsupervised classification of the given inputs. Hence, we can say that this paper proposes a new method to build a hierarchical classifier, (with a mathematical basis) a new kind of common algorithm (which is implemented throughout the hierarchy of the classifier) and it is demonstration on an example problem.

We find it necessary to discuss a few points about the MNN before moving on to the details of the proposed Theorem and the Tandem Algorithm. An MNN is nothing but a neural network (NN) with converging-diverging architecture which is trained to produce an output which equals its input as closely as possible (i.e. mirror the input at its output layer). And this training process proceeds through repeated presentations of all the input samples and it stops when the MNN could mirror at least above 95% of its input samples. Then the MNN said to be successfully trained with the given input samples. Now, the best possible extracted features of the inputs are automatically obtained at the MNN's least dimensional hidden layer and these features are used for unsupervised input classification by a clustering algorithm. See Figure 1 for illustration of an MNN architecture wherein input given to it is 'X' of dimension 'n' which is reduced to 'Y' of dimension 'm' (m is much less than n). Since Y is capable of mirroring X at the output, Y contains as much information as X, even though it has a lesser dimension, the components of Y can then be thought of as features that contains the patterns in X, hence, the Y can be used for classification. More details on MNN architecture can be referred from [20] & [21].

Before, proceeding to proving the main theorem and the presentation of actual computer simulation, it is perhaps appropriate to write a few lines on the ideas that motivated this paper.

It is presently well known that the neural architecture in the human neocortex is hierarchical [22], [23], [24] & [25] and constituted by neurons at different levels and information is exchanged between these levels via these neurons [26], [27], [28] & [29] when initiated by the receipt of data coming in from sensory receptors in the eyes, ears, etc. The organization of the various regions within each level of the neo-cortical system, are not completely understood, but there is much evidence that regions of neurons in one level are connected with regions of neurons in another level thus forming many tree like structures [25] & [30] (also see [31]). Various intelligent tasks, for example "image recognition", are performed by numerous neurons firing and sending signals back and forth these levels [32]. Many researchers working in the field of artificial intelligence have sought to imitate the human brain in their attempt to build learning machines [33] & [34] and have employed a tree like architecture at different levels for performing recognition tasks [35]. As described above, our attempt here is to demonstrate that a hierarchical classifier which addresses the tasks of feature extraction (/data reduction) and recognition can be constructed and such architecture can perform intelligent recognition tasks in an unsupervised manner.

The plan of the paper is as follows: In the next section we prove the proposed Mirroring Theorem of Pattern Recognition. In section 3, based on the proof of the mirroring theorem, we show how to build pattern classifiers which possess the ability to automatically extract features, have a tree-like architecture and can be used to develop the proposed architecture for unsupervised pattern learning and classification (including the proposed tandem algorithm). In section 4, we report the results of the demonstration of such a classifier when applied an unsupervised pattern recognition problem wherein real images of faces, flowers and furniture are automatically classified and sub-classified in an unsupervised manner. Section 5, we discuss the future possibilities of this kind of architecture.

## II. MIRRORING THEOREM

We now prove what we term as the mirroring theorem of pattern recognition,

**Statement of the Theorem:**

If a hierarchical invertible map exists that

(i) maps a set $\wp$ of $n$-dimensional from $X$-space into a $m$-dimensional data in $Y$-space ($m \leq n$) which fall into $j$ distinct classes, and,

(ii) if for each of the above j classes, in turn, maps exist which map each class in $Y$-space to a $r$-dimensional $Z$-space into $t$ subclasses, then such a map can be approximated by a set of $j + 1$ nodes (each of which are MNNs with an associated clustering algorithm) forming a treelike structure.

**Proof:**

The very fact that invertible maps exist indicate that there exist $j$ 'vector' functions which map the points $(x_1, x_2, x_3,...x_n)$ falling into some $d$ different regions $R_d$ in $Y$-space. These 'vector' functions may be denoted as: $\underline{F}^1$, $\underline{F}^2$,....., $\underline{F}^j$. We clarify our notation here by cautioning that $\underline{F}^1$, $\underline{F}^2$,....., $\underline{F}^j$, should <u>not</u> be treated as the vectoral components of $\underline{F}$. What we mean by $\underline{F}^1$ are the maps that carry those points in $X$-space to the set of points contained in $S_1$, hence $\underline{F}^1$ can be thought of as a collection of 'rays', where each 'ray' denotes a map starting from some point in $X$-space and culminating in a point belonging to $S_1$ in $Y$-space. Similarly, $\underline{F}^2$ is the collection of 'rays' each of which starts from some point in $X$-space and culminates in a point belonging to $S_2$ in $Y$-space. Thus we define the map $\underline{F}$ as $\underline{F} \equiv \underline{F}^1 \; U \; \underline{F}^2 \; U......U \; \underline{F}^j$.

Now we argue as follows: since the first map $\underline{F}^1$ takes $X$-space into an image in $Y$-space, say a set of points $S_1$ and similarly $\underline{F}^2$ takes $X$-space into an image in $Y$-space, say a set of points $S_2$ and so to the set $S_j$ and since, by assumption, the target (image) region in $Y$-space contains $j$ distinct classes, we could conclude that the set of points $S_1$, $S_2$, ..., $S_j$ are distinct and non overlapping, (for otherwise the assumption of there being $j$ distinct classes is violated). These regions are separable are distinct from one another and there also exist maps that are all distinct, and we can renumber the regions $R_d$ in such a manner that the union of the first $k_1$ sets belong to $S_1$ i.e., $S_1 = R_1 U R_2 U....U R_{k1}$ and the union of the next $k_2$ sets belong to $S_2$ ..... and similarly $S_2 = R_{k1+1} U R_{k1+2} U.....U R_{k1+k2}$ , e.t.c., till $S_j = R_{d-kj+1} U R_{d-kj+2} U....U R_d$. It also implies, since each of the image sets can be again reclassified into k patterns (by assumption), that there exists a 'vector' function set $\underline{G_l}^1$, $\underline{G_l}^2$, $\underline{G_l}^3$ , ..., $\underline{G_l}^t$ which take $S_1$ to $t$ distinct regions in z-space, these $t$ distinct sets are denoted by $c_{11}$, $c_{12}$, $c_{13}$, ...., $c_{1t}$. Again, here $\underline{G_l}^1$, $\underline{G_l}^2$, $\underline{G_l}^3$ , ..., $\underline{G_l}^t$ can be thought of a collection of 'rays' denoting maps from points in $S_1$ in $Y$ -space to points in the sets $c_{11}$, $c_{12}$, $c_{13}$, ...., $c_{1t}$ in Z-space. Thus $\underline{G_l}^1$ is the collection of 'rays' leading to the set $c_{11}$ from $S_1$ and $\underline{G_l}^2$ is the collection of 'rays' from $S_1$ to the set $c_{12}$. Hence, similar to the definition of $\underline{F}$ we can denote the map $\underline{G_l}$ as $\underline{G_l} \equiv \underline{G_l}^1 \cup \underline{G_l}^2 \cup \underline{G_l}^3 \cup ..... \cup \underline{G_l}^t$. In order not to clutter the diagram the possible sub-regions within each of the sets $c_{11}$, $c_{12}$, $c_{13}$, ...., $c_{1t}$ have not been drawn in Figure 2 and we assume, without prejudicing the theorems generality, that the number of subsets $t$ are the same in all maps.

Similarly $\underline{G_2}^1$, $\underline{G_2}^2$, $\underline{G_2}^3$, ..., $\underline{G_2}^t$ take $S_2$ to distinct $t$ sets $c_{21}$, $c_{22}$, $c_{23}$, ..., $c_{2t}$ and so on to the function set $\underline{G_j}^1$ , $\underline{G_j}^2$, $\underline{G_j}^3$ , ..., $\underline{G_j}^t$ which map $S_j$ to respective $c_{j1}$, $c_{j2}$, $c_{j3}$,..., $c_{jt}$ in Z-space.

The existence of the function maps $\underline{F}^1$, $\underline{F}^2$,...., $\underline{F}^j$ which map points from the set $\wp$ in the $n$ dimensional space to j distinct classes implies that the set of points in $\wp$ are separable to j distinct classes in Y-space which is of $m$ dimensions. (Strictly speaking it is necessary to assume that these functions $\underline{F}^i$, $i = 1, 2, ...., j$ have the property of being invertible (i.e., are bijective) in order to exclude many-to-one maps; also this property is necessary to prove that the function such as $\underline{F}$ can be approximated by an MNN along with a clustering algorithm. Further, it is also being implicitly assumed that all maps considered in this theorem are at least continuous up to first order, points which are close to one another are expected to have their images close to one another).

To proceed with the proof we will first show that it is possible to approximate the set of maps that take $X$-space to $Y$ - space by a single MNN. To do this we will show an MNN can explicitly be constructed and trained to perform this map. We will assume that sufficient samples are available for training. Now consider the MNN to have a converging set of layers of adalines (PE's), the converging set consists of '$n$' inputs in the first layer and ends with a layer of '$m$' adalines, shown in Figure 1(a).

This set from '$n$' to '$m$' can be thought as a simple NN which can be trained such that if $\underline{X} = (x_1, x_2, x_3,...x_n)$ is the input then $\underline{Y} = (y_1, y_2, y_3,...y_m)$ is the output, the weights of this network can be obtained by using the back propagation algorithm to impose the following conditions on the output: $\underline{Y} = \underline{F}^k (x_1, x_2, x_3,...x_n)$ where $k$ is the class to which the input vector $(x_1, x_2, x_3,...x_n)$ belongs obviously $k$ is known before hand because the $\underline{F}$'s are known. Thus we can train this converging part of the NN.

Similarly, we can now assume that there exists a diverging NN (depicted in Figure 1 (b)) to exist starting from '$m$' adalines and ending in '$n$' adalines, to this second neural network we will assume that the input will be the set $(y_1, y_2, y_3,...y_m)$ and the output of this would be the original point in $X$ dimension space whose image is $Y$. So by imposing the latter condition the second (diverging neural network) can be trained with sufficient samples and then the weights obtained. At this stage we have a diverging neural network which takes as input $\underline{Y}$ and outputs the corresponding $\underline{X}$. Now by combining the two converging and diverging so that the first leads to the second (without changing the weights) we have nothing but an MNN (pictorially represented by Figure 1(c)), this MNN mirrors the input vector $\underline{X}$ and outputs $\underline{Y}$ from the middle layer of '$m$' adalines. So, we have thus proved the existence of an MNN which maps points from the $n$-dimensional X space to the $m$-dimensional Y space and then back to the original points in $n$-dimensional X space. Then the points in Y space can be classified into j classes using a suitable clustering algorithm. Thus, we have proved that a node of the hierarchical classifier is approximated by the combination of an MNN and a clustering algorithm.

The proof that the second set of maps from $m$ space to $r$ space exists, uses similar arguments. We can immediately see that there will be $j$ maps because there are $j$ classes in Y space, hence there will be $j$ nodes for each class each of which constructed by using a similar procedure. So we see that the set of maps assumed can be approximated by $j + 1$ nodes, whose existence we have just proved, all of which forming a treelike structure, shown in Figure 3 QED. It may be noted that each node in our architecture is depicted in Figure 4.

We will now illustrate, the use of the mirroring theorem to develop a hierarchical classifier which performs the task of unsupervised pattern recognition.

## III. UNSUPERVISED HIERARCHICAL PATTERN RECOGNITION

This section describes the architecture of a self-learning engine and the next section, we report its application to an example problem, wherein a set of input images are automatically classified and then sub-classified.

Our intent is to build a learning engine which has the following characteristics: It is (i) hierarchical (ii) modular (iii) unsupervised and (iv) runs on a single common algorithm (MNN associated with clustering). The advantage of developing a recognition system with these 4 characteristics is that the learning method does not depend on the problem size and the learning network can be simply extended as the recognition task becomes more complex. It has been surmised by investigators that the architecture of human neo-cortex does, loosely speaking, possess the above 4 characteristics (except that instead of (iv) there is some kind of analog classification process (procedure) performed by sets of neurons, which seemingly behave in a similar manner). We are also reinforced by the conviction, since our architecture imitates the neural architecture (though admittedly in a crude manner), it is reasonable to expect that we would meet with greater successes as we make improvements in our techniques and as we deal with problems of larger size using increasingly powerful computers. In fact, it is this prospect that has been the prime motive force behind our work.

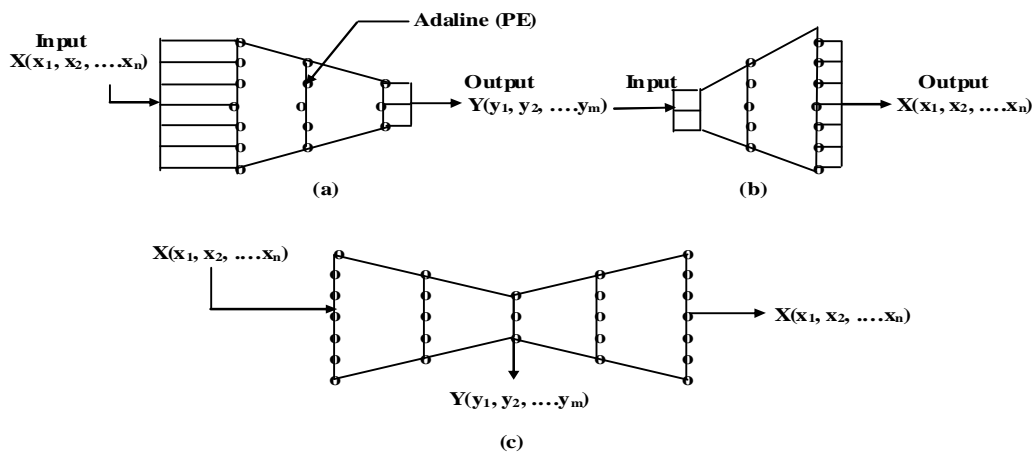

Figure 1. (a) Convergning NN (b) Diverging NN (c) Mirroring Neural Network (combining (a) and (b))

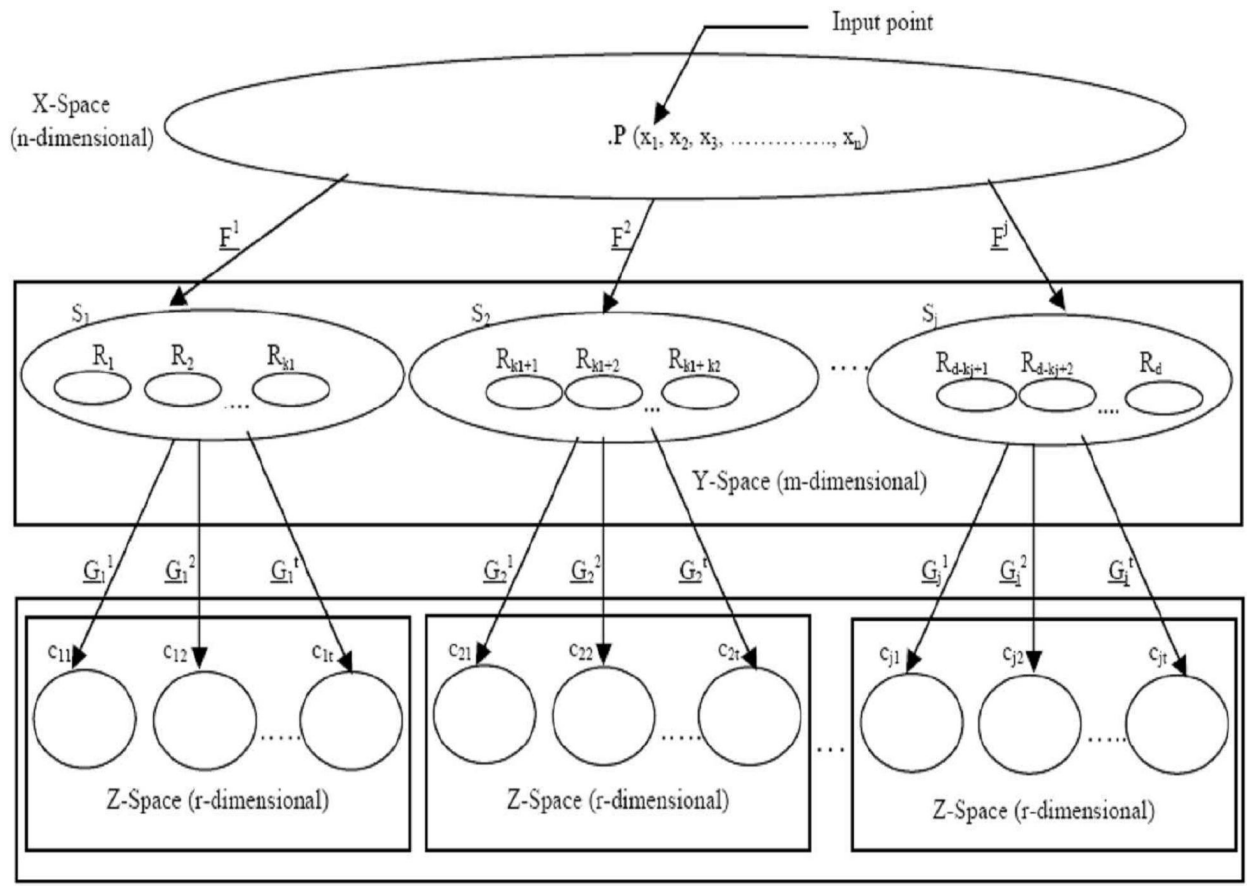

Figure 2.  An Illustration of the hierarchical invertible map

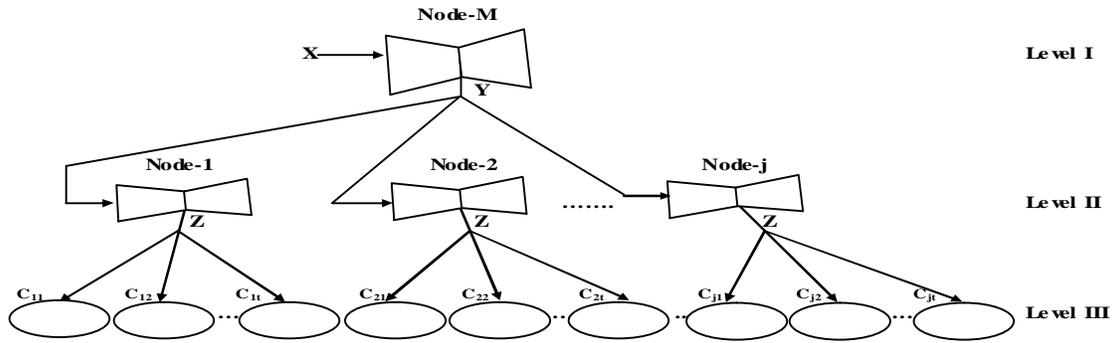

Figure 3. Organized collection of Nodes (blocks) containing MNN's and their corresponding Forgy's algorithm – Forming a treelike hierarchical structure

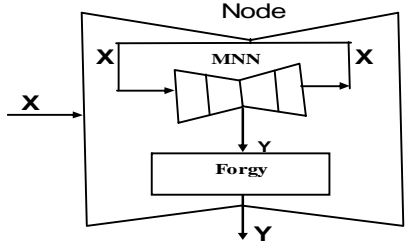

Figure 4. A Node (block) of the Hierarchical Classifier constructed with MNN and Forgy's clustering

The Tandem Algorithm which we devised in this paper for pattern recognition tasks using a hierarchical tree-like architecture (depicted in Figure 3). It may be noted that each block in the hierarchical architecture is trained through the implementation of a single common algorithm (tandem algorithm). This tandem process is done (at each node) in two steps. The 1st step being the data reduction and feature extraction by an MNN and the 2nd step is the classification of the extracted features (outputs of the MNN) using a clustering algorithm. The MNN at the first level trains itself to extract the features through repeated presentations of the data samples, after which the extracted features of the data are sent to the clustering procedure for classification. The modules in the second level again undergo this tandem process of feature extraction and classification (/sub-classification). This is how a single common algorithm is implemented throughout the hierarchy (at each module), resulting a level-by-level unsupervised classification. In section 4, we show that our method actually works: we apply our classifier on a collage of images of flowers, faces and furniture, this collection is automatically classified and sub classified.

We will now develop the tandem algorithm and actually implement it by writing a computer program by which such a learning engine can be used to classify the patterns by itself and report the results. The technique used for the development of this algorithm is based upon the application of the two procedures (i) mirroring for automatic feature extraction and (ii) unsupervised classification, in a tandem manner as described by the following algorithm, at each module (block) of the hierarchy, level-by-level. (In our computer program we have used it on a two level hierarchy). Continuing the discussion of the Tandem Algorithm, consider Figure 3 which is a pictorial representation of the hierarchical architecture, the details of each block or node is shown in Figure 4 and the structure of a MNN in Figure 1. The tandem Algorithm proceeds block (node) by block (node) at each level starting from the 1st level (see Figure 3).

**The Tandem Algorithm for a hierarchical classifier:**

1. Train the MNN of the topmost block, i.e. Node-M (of the hierarchy, see Figure 3) with an ensemble of samples such that the inputs given to the MNN are mirrored at its output with minimum loss of information (for which a suitable threshold is fixed). And mark the topmost node as the "present node". This is an iterative step and stops when the output of Figure 1c, almost equals the input, that is able to reconstruct the input.

2. After due training of the MNN of the present node (i.e., the MNN could accurately reconstruct above 95% of its inputs within the specified threshold limit), the collection of outputs of the MNN's least dimensional hidden layer (the extracted number of features is equal to the dimension of Y of the MNN see Figure 1c) is considered for classifying the input of the present node.

3. The features extracted in step 2 are given as "input data set" to the Forgy's clustering algorithm (subroutine) of the present node for unsupervised classification, explained in step 4.

4. The steps involved in clustering procedure are:

   a. Select initial seed points (as many in number as the no. of classes the inputs to be divided into) from "input data set".

b. For all the data samples in the input dataset repeat this step b.

(i) Calculate distance between each sample in the input data set and the each of the seed points representing a cluster.

(ii) Place the input data sample into the group associated with the seed point which is closest (least of the distances in step 4 b (i)).

c. When all the samples are clustered, the centroids of each of the clusters considered as new seed points.

d. Repeat step 4 b, 4 c as long as the data sets leave one cluster to join another in step 4 b (ii).

5. To proceed further for sub-classification, repeat step 6 for all the nodes in the next below level of the hierarchy.

6. Mark one of nodes as the "present node" and train the MNN of the present node with the samples (extracted features of the immediate above level) belong to a particular cluster of step 4 such that the samples given to the MNN are mirrored at its output with minimum loss of information (for which a threshold is fixed by trail and error). Repeat steps 2, 3 and 4.

7. Repeat steps 5 and 6 till there is no enough data present in the samples to be further sub-classified (at the immediate below level).

In this tandem algorithm, the feature extraction (concurrent with data reduction) is through steps 1, 2 and 3 and the automatic data classification (based on the reduced units of data) is by step 4. This tandem process of data reduction and classification is extended to next lower levels of the hierarchy for sub-classifying the ensemble through steps 5 and 6 till the stated condition is met in step 7.

More details on the MNN architecture and MNN's training through self-learning are given in [20] & [21].

We now, illustrate this concept of hierarchical architecture for unsupervised classification using Figure 3. If we assume, for the purpose of illustration, that there are only 4 categories of images; say faces, flowers, furniture and trees (j = 4), then at its broadest level, the MNN-M at Node-M is trained with these 4 categories of images. On successful training, MNN-M can reduce the amount of its input data; and based on the reduced units of data, Node-M categorizes the pattern into one of the classes using Forgy's algorithm. The reduced units (which represent the input data) of the pattern from the present node (Node-M) are fed to one of the next level (Level II) nodes. (Alternatively, the input vector could be fed to the appropriate MNN in next level (Level II), instead of the reduced vector, in cases where too large an amount of data reduction done at the present level (Level I), is expected to have loss of information required for the finer classification at Level II). Selection of a node (module) from next level depends upon the classification of the input pattern at the present level. For example, Node-1 is selected if Node-M classifies the input as a face, else Node-2 is selected if Node-M classifies the same input as a flower and so on for Node-3 (furniture) or Node-4 (tree). Then, the respective node

(module) at Level II reduces its input and does a sub-classification (we denote it as Level II classification) based only on its reduced units (at Level II). The gender classification which distinguishes a male face from a female face is a typical Level II classification by Node-1. In the pictorial representation, Level II classification contains 't' subcategories in each of j categories. Assuming that there are some more lower levels (identical to Level I and/or Level II) containing the nodes to further classify the patterns, so, for instance, the reduced units at Level II are given as input to one of the appropriate modules at Level III for more detailed classification which, an example case, sub-categorizes 'k' different persons in male/female face group. This tandem procedure of (i) mirroring followed by (ii) classification, performed at each level, can be extended to other lower levels, say, level IV, V and so on. That is how; the proposed architecture does level-by-level unsupervised pattern learning and recognition.

As explained earlier, the hierarchical architecture implements a common algorithm for data reduction and extracted feature classification at its each node. And as the data reduction precedes the classification, the accuracy of classification is dependent on the efficiency of the data reduction algorithm. So there is a need to evaluate the performance of the MNN's data reduction. The fact that the MNN dimensional reduction technique is an efficient method to reduce the irrelevant parts of the data was amply demonstrated over extensive trials (details are in [20] & [21]). It is because of this that we used the MNN (along with clustering algorithm) as a data reduction and feature extraction tool for the hierarchical pattern classification. For our demonstration, we use the Forgy's algorithm for clustering the reduced units (of the input, at each module), wherein the number of clusters for the classification/sub-classification is provided by the user. Instead, without prejudice to the generality of our technique, one could use a more sophisticated clustering algorithm wherein the number of classes (clusters) is determined by the algorithm. We leave this work as a part of future enhancement which would then result in a completely automated unsupervised classification algorithm.

## IV. DEMONSTRATION AND RESULTS

We now show by explicit construction that a hierarchical architecture can actually be built and used for classification and sub-classification of images, giving an example case.

**Example:** Here we took a collection of 360 images for training with an equal no. of faces (See databases Feret [36], Manchester [37], Jaffe [38] in references), tables and flowers. We build a two level classifier constructed out of MNNs (associated with Forgy's clustering); which at the first level automatically classifies the 360 images of the training set into three classes one of them would be a "face class" and the other two belong to the "table class" and "flower class". The automatic procedure which does this is as follows: A 4 layer MNN (676-60-47-676) consisting an input layer of 676 inputs representing a 26 X 26 image, with the 60 processing elements in the first layer and 47 and 676 processing elements in the other two layers is used to train the MNN to mirror the input data on to itself. The training is done automatically and stops when the output vector of 676 dimensions closely matches the corresponding input vector for each image, at this point, the MNN can said to be satisfactorily mirror all the 360 input images.

Then the output of the layer with the least number of processing elements (in this case 47) is taken as a reduced feature vector for the corresponding input image. We would have a set of 360 vectors (each of 47 dimensions) representing the input data of 360 images. This set of 360 vectors (of reduced dimensions) is then classified into three classes by using Forgy's Algorithm (see [39] & [40]). The actual classification varies somewhat on the choice of the initial seed points which are randomly chosen. The program chooses three distinct initial random seed points and uses Forgy's algorithm to cluster the reduced vectors of the input images. This clustering is done over many iterations till convergence and the classes are then frozen; after this the data is clustered a second time (starting from the second set of seed points) again using Forgy's algorithm till another set of three classes are obtained. After this the average of the resulting two nearest sets of cluster centroids is considered as the new cluster centroid, based on which the reduced feature vectors are once again classified to obtain three distinct classes, these classes are then treated as the final three classes (if everything works out well one of them would be the face class and the other remaining two  would be the table class and flower class).

After this first level classification, the program proceeds to the next level for sub-classifying the three classes identified at level I. The procedure of reduction and classification at this Level II, is similar to that carried out at Level I, except that now three MNNs have to be trained, one receiving inputs form the Face class, another from Table class and the other from Flower class. These MNNs at Level II use the architecture (47-37-30-47). After the two MNNs are suitably trained to mirror their respective inputs, to an acceptable accuracy, the program proceeds to classify the inputs into sub categories for each of the MNNs separately. Of course, this time the feature vector (reduced vectors) has 30 dimensions. Once again, Forgy's Algorithm is used, following a similar procedure as described above for level I, except that this time the classification is done on the reduced vectors of the MNN-1 at Node-1 which would render the sub categories male face and female face, a classification of the reduced vectors of the MNN-2 at Node-2 obtaining the subcategories centrally supported table and four legged table and a classification of the reduced vectors of the MNN-3 at Node-3 obtaining the subcategories flower bud  and open flower.

Because the MNNs are initiated with random weights (chosen initially), and again by choosing random seed points while executing the Forgys Algorithm, it is our intention to demonstrate that the classification is not overly dependent on these random choices. So, we ran the program over and over again each time starting ab initio. We have taken 10 trials, meaning, 10 different training and classification sessions at level I followed by level II. On an average of these 10 trails, considering the training and test sets, the error  at level I  is 7% and an average error of the three nodes at level II (for subcategorizing a "face" as "male" or "female", a "table" as "centrally supported" or "four-legged" and a "flower" as "flower bud" or "open  flower") is an additional 7%. Actually, this is not too bad at all because the whole exercise is unsupervised and the errors made in the 1$^{st}$ level classification remain undiscovered and are actually uncorrected by the classifier which indiscriminately feeds all the data into the second level as inputs.

See the sample illustration for the Example in Figure 5. The summary of the results for Example is given in Table I. The various parameters used in the MNN training and classification are given in Table II.

The brute force (obvious procedure) of training the MNN at each node of the hierarchical classifier by using a Newton-Raphston is beyond the capability of the PCs available with us and was not tried. Instead, we adopted an approximate procedure and trained the MNNs by using the Back Propagation algorithm ([41] & [42]) which actually tries to determine the best MNN by changing the weights at each presentation of an image; ideally a "best MNN" should be obtained for the entire ensemble of input images (or reduced units of images) at each MNN of a node, which again would involve a Newton-Raphston and was avoided. The techniques used and reported here were very efficient in terms of time and space taken for execution and they were all performed on a PC.

## V.  SIGNIFICANCE OF OUR WORK & CONCLUSIONS

In this paper we have proved a crucial theorem called "Mirroring Theorem"; based on the mathematical proof of the theorem we developed an architecture for a hierarchical classifier which implements our proposed Tandem Algorithm to perform an unsupervised pattern recognition. We have also specifically written a computer code to demonstrate the technique of building such a self-supervising classifier and applied it for an example. These classifiers have the characteristics of being hierarchical, modular, unsupervised and they run on a single common algorithm and therefore, they mimic (admittedly in a crude manner), the collective behavior of neurons in the neo-cortex. It is expected that they can be expanded to analyze much more complex data, such "super classifiers" could employ many structures, (each being of the type shown in Figure 3), working in parallel.

In our experimentation, (within the available resources) we have found that it is not possible to have too many classes at the first level (Figure 3), i.e. j cannot be too large a value (at best j = 4). Therefore,  for large problems involving many classes, we need to have a network of "structures" (each being of the type shown in Figure 3 but with j limited to 2, 3 or 4) working in parallel, each structure trained to recognize its own set of classes (eg. face classes, alphabet classes etc.). Thus a binary or tertiary "super- tree" with each "node" itself being a structure of type shown in Figure 3, can be envisaged for the construction of a "super classifier".

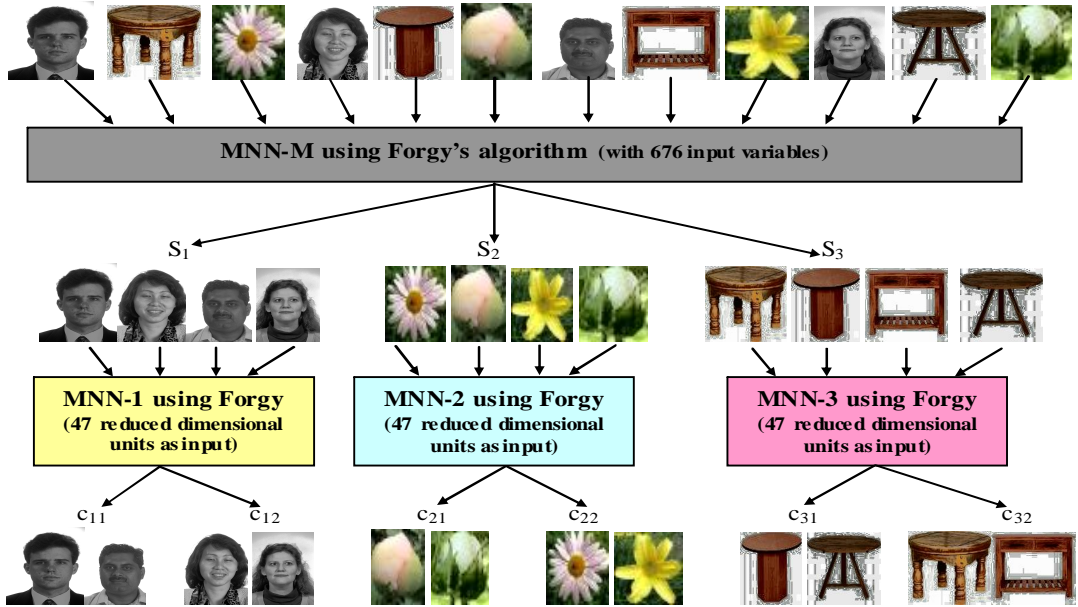

Figure 5  Pictorial representation of Hierarchical classifier implemented using Example images;
($S_1$ (face), $S_2$ (flower), $S_3$ (table): classification at level I based on 47 reduced dimensional vectors of the input image;
$c_{11}$ (male face), $c_{12}$ (female face), $c_{21}$ (flower bud), $c_{22}$ (open flower), $c_{31}$ (centrally supported table), $c_{32}$ (four-legged table): sub-classification at level II based on 30 reduced dimensional vectors of the image).

TABLE I.    RESULTS OF THE HIERARCHICAL CLASSIFIER FOR EXAMPLE IMAGES

| Input type | Dimension of the input | Dimension of the reduced units | No. of samples for training | No. of samples for testing | No. of categories | Success rate (averaged over 10 trails) of clustering on reduced units | |
|---|---|---|---|---|---|---|---|
| | | | | | | *Training samples* | *Average of Training & Test sets* |
| **Image** | 676 (26 X 26 ) | 47 | 360 | 150 | 3(face,  table & flower) | 94.0% (Efficiency of the Level I Node) | 93.4% (Efficiency of the Level I Node) |
| **Reduced units of image** | 47 | 30 | ≈ 120 (for each category) | ≈ 50 (for each category) | 2 (sub-categories for each category) | 88.5%(Average efficiency of the level II  Nodes) | 86.3%(Average efficiency of the level II Nodes) |

TABLE II.    VARIOUS PARAMETERS USED FOR THE MNN AND FORGY'S ALGORITHM

| Type of MNN architecture | Distance between input and output | Seed points for Forgy's algorithm | Learning rate parameter | Weights& bias terms |
|---|---|---|---|---|
| Level I MNN (676-60-47-676) | 0.8 | Threshold of 1.0, between the random seed points | 0.025 | -0.25 to +0.25 (random selection) |
| Level II MNNs (47-37-30-47) | 0.8 | Threshold of 0.8, between any two random seed points | 0.01 | -0.25 to +0.25 (random selection) |

It is expected that the techniques that we have developed and presented in this paper will be implemented by many future researchers for building advanced and highly sophisticated pattern classifiers. Further it is also hoped that these procedures will also be used for building models for associative memories [43] where, say a voice signal (eg. "Mary": a spoken word) can be associated with a picture (image of Mary). These developments could, in the near future, lead to very versatile machine learning systems which can possibly ape the human brain in at least its elemental functions.

ACKNOWLEDGMENT

We thank the managements of Srinidhi and the group of Aurora Educational Institutions for their encouragement and Altech Imaging and Computing for providing the facilities for research.

## AUTHORS PROFILE

**Author 1:** Working as an Associate Professor (CSE Dept.), Aurora's Engineering College. Submitted Ph. D.(CSE) Thesis on "Automatic pattern recognition for applications in image processing and robotics" to Osmania University, Hyderabad in  Feb. 2009. M. Tech. (Software Engineering) from J. N. T. University, Hyderabad.

**Author 2:** Working as a Professor (CSE Dept.), Srinidhi Institute of Science Technology.  Ph. D. (Mathematical Physics) on "On Phase and Coherence in Quantum Systems" from University of Madras, Jan. 1973. 36 years of research experience in the application of Computers in the areas of Industrial Image Processing, Pattern Recognition, Neural Networks, Electromagnetics, Fluid Mechanics, Structural Mechanics and Artificial Intelligence. He has more than 40 papers in international journals and international conferences on the above subjects.